\documentclass[letterpaper, 10pt, conference]{ieeeconf} 
\IEEEoverridecommandlockouts     
\usepackage{graphics} 
\usepackage{graphicx}
\usepackage{caption}
\usepackage{subcaption}
\usepackage{algpseudocode}
\usepackage{algorithm}
\usepackage[]{units}
\usepackage{paralist}
\usepackage{mathtools}

\usepackage{xcolor}
\usepackage{siunitx}
\usepackage{multirow}
\usepackage{epstopdf}
\graphicspath{ {Pictures/} }
\usepackage{multicol}
\usepackage{epstopdf}
\usepackage{tikz}
\usetikzlibrary{shapes,arrows}
\usepackage{epsfig}
\usepackage{float}
\usepackage{caption}
\usepackage{amsmath} 
\usepackage{amssymb}  
\usepackage{mathtools}
\graphicspath{ {Pictures/} }
\DeclareGraphicsExtensions{.pdf,.eps}
\title{\LARGE \bf Cooperative Aerial Coverage Path Planning \\for Visual Inspection of Complex Infrastructures
\author{Sina Sharif Mansouri, Christoforos Kanellakis, David Wuthier, Emil Fresk and George Nikolakopoulos
\thanks{The authors are with the Robotics Group at the Control Engineering Division of the Department of Computer, Electrical and Space Engineering, Lule\r{a} University of Technology, Lule\r{a} SE-97187, Sweden}
\thanks{This work has received funding from the European Unions Horizon 2020 Research and Innovation Programme under the Grant Agreement No.644128, AEROWORKS.}}}
\begin{document}
\maketitle
\thispagestyle{empty}

\begin{abstract} 
This article addresses the problem of Cooperative Coverage Path Planning (C-CPP) for the inspection of complex infrastructures (offline 3D reconstruction) by utilizing multiple Unmanned Autonomous Vehicles (UAVs). The proposed scheme, based on a priori 3D model of the infrastructure under inspection, is able to generate multiple paths for UAVs in order to achieve a complete cooperative coverage in a short time. Initially the infrastructure under inspection is being sliced by horizontal planes, which has the capability of recognizing the branches of the structure and these branches will be handled as breaking points for the path planning of the UAVs to collaboratively execute the coverage task in less time and more realistically, based on the current flying times of the UAVs. The multiple data sets collected from the coverage are merged for the offline sparse and dense 3D reconstruction of the infrastructure by utilizing SLAM and Structure from Motion approaches, with either monocular or stereo sensors. The performance of the proposed C-CPP has been experimentally evaluated in multiple indoor and realistic outdoor infrastructure inspection experiments. 
\end{abstract}
%
\section{Introduction}
%
The annual investments on the infrastructure sector represent a significant percentage of the Gross Domestic Product (GDP) of developed and developing countries e.g. 3.9\% of the GDP for the old European states, 5.07\% of the GDP for the new European states and 9\% of the GDP for China~\cite{EuropeanInvestment}. In order to decrease the human life risk and to increase the performance of the overall procedure, autonomous ground, aerial or maritime vehicles are employed for executing the inspection tasks. As an example, for these applications it can be mentioned the power-line monitoring using autonomous mobile robots~\cite{fernandes1990line}, bridge inspection~\cite{ metni2007uav}, boiler power-plant 3D reconstruction~\cite{burri2012aerial}, urban structure coverage~\cite{cheng2008time}, forest fire inspection~\cite{alexis_coordination_2009} using UAVs, and inspection of underwater structures or ship hulls as in~\cite{galcerancoverage} and~\cite{englot2012sampling} respectively by the utilization of  autonomous underwater vehicles. In most of these scenarios, there is an a priori knowledge about the infrastructure, while the 3D or 2D models are available or can be derived using CAD software.

In general, the task of Coverage Path Planning (CPP)~\cite{galceran2013survey} has received significant attention over the last years, however still there are limited CPP approaches in the case of aerial robotics. In the case that the CPP concept is extended in the Collaborative approach (C-CPP) by the utilization of multiple aerial agents instead of single one, the overall objective of the coverage time has the potential to be dramatically reduced, while it can be achieved realistically by multiple UAVs, when taken under consideration the flying times and the levels of autonomy. Thus, inspired by this vision, the main objective of this article is to establish a C-CPP method that is based on an a priori knowledge of the infrastructure (e.g. a CAD model) and it will have the ability to generate proper way points by considering multiple agents, while guaranteeing the coverage objective and the overall collision avoidance among the flying agents. As it will be presented, the scheme has the capability to detect branches of complicated infrastructures and denote these branches as breaking points, which are the points that create sub-coverage path planning for cooperative inspection of the whole infrastructure. 

More specifically, in the related literature there have been many works that addressed the CPP problem in 2D spaces and fewer approaches that address coverage of 3D spaces~\cite{galceran2013planning}, while in~\cite{galceran2013survey} a complete survey was presented on CPP methods in 2D and 3D. Towards 3D CPP, Atkar et al.~\cite{atkar2001exact} presented an offline 3D CPP method for spray-painting of automotive parts. Their method used CAD model and the resulting CPP should satisfy certain requirements for paint decomposition. In~\cite{galceran2014coverage}, the author presented a CPP with real time re-planning for inspection of 3D underwater structures, where the planning assumed an a prior knowledge of a bathymetric map and they adapted their methodology for the case of an autonomous underwater vehicle, while their overall approach was containing no branches. The authors in \cite{englot2012sampling} introduced a new algorithm for producing paths that cover complex 3D environments. In this case, the algorithm was based on off-line sampling with the application of autonomous ship hull inspection, while the presented algorithm was able to generate paths for structures with unprecedented complexity.  

In the area of using UAVs for inspection, \cite{cheng2008time} presented a time-optimal UAV trajectory planning for 3D urban structure coverage. In this approach, initially the structures to be covered (buildings) were simplified into hemispheres and cylinders and in a later stage the trajectories were planned to cover these simple surfaces. In~\cite{bircher2015three}, the authors studied the problem of 3D CPP via viewpoint resampling and tour optimization for aerial robots. More specifically, they presented a method that supports the integration of multiple sensors with different fields of view and considered the motion constraints on aerial robots. Moreover, in the area of multi-robot coverage for aerial robotics in~\cite{barrientos2011aerial}, a coverage algorithm with multiple UAVs for remote sensing in agriculture has been proposed, where the target area was partitioned into $k$ non-overlapping sub-tasks and in order to avoid collision both different altitudes have been assigned to each UAV and security zones were defined where the vehicles are not allowed to enter. 

Based on aforementioned state of the art, the main contribution of this article is double. Firstly, to automate the inspection of complex 3D structures with multiple agents and reduce the inspection time, the concept of cooperative aerial inspection is being experimentally verified and demonstrated within the novel proposed scheme for infrastructure inspection. In this novel approach, the a priori coverage path is divided and assigned to each agent based on the infrastructure architectural characteristics. Furthermore, to guarantee full coverage and 3D reconstruction, the introduced path planning for each agent create an overlapping visual inspection area, that will enable the off line cooperative reconstruction. In the novel established C-CPP scheme, the introduced algorithms take into account the non-convex structure and identify its branches. The algorithm, additionally to the position references, provides yaw references for each agent for assuring field of view directed to the structure surface. 
\begin{figure}[H]
    \centering
        \includegraphics[width=40mm]{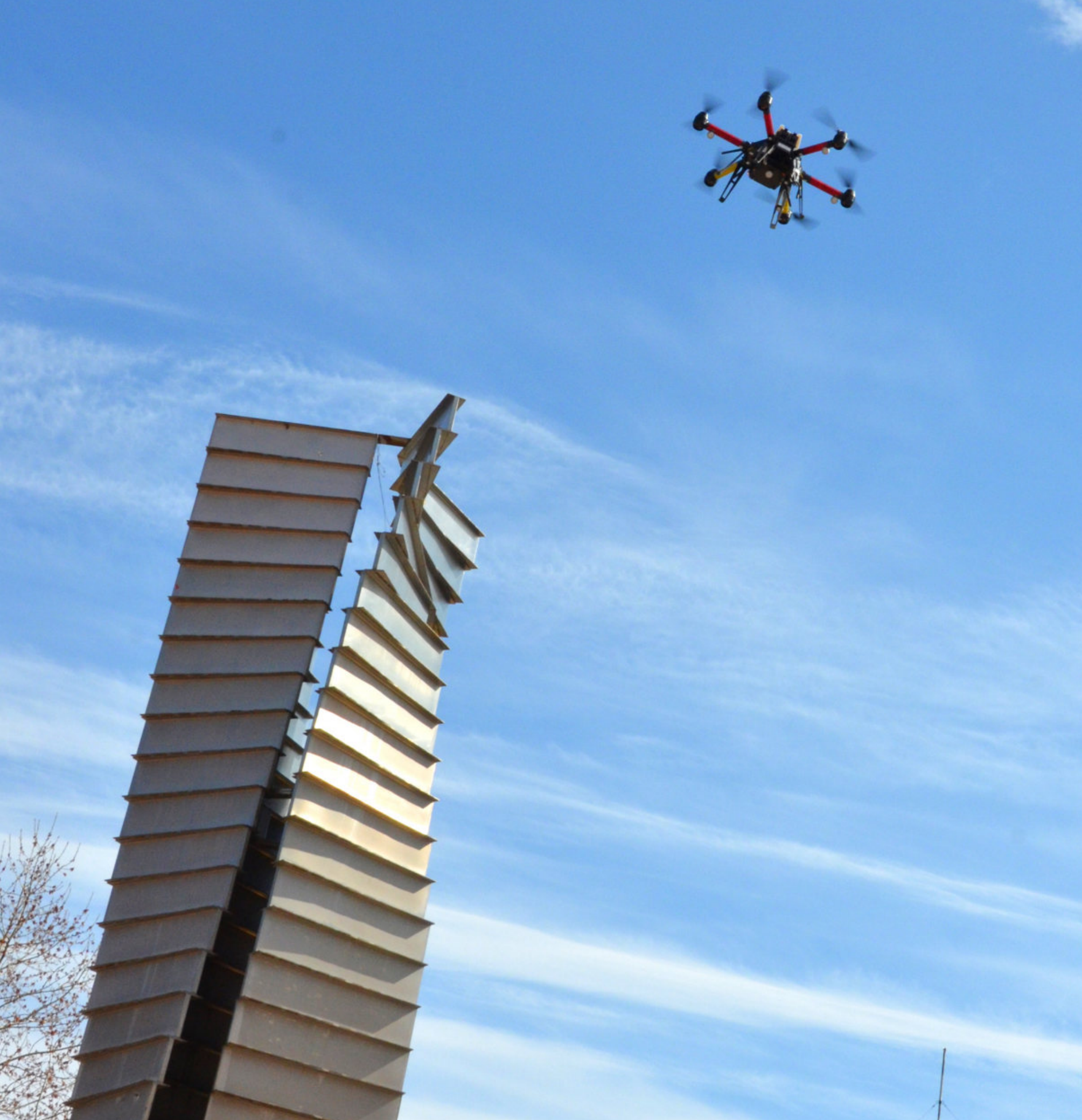}
        \caption{Hexacopter during outdoors coverage task.}
        \label{fig:hexacopter_cov}
    \end{figure}
The second major contribution stems from the direct demonstration of the applicability and feasibility of the overall novel C-CPP scheme for both indoors (simple structure) and outdoors (complex structure). This demonstration has a significant novelty and impact as an enabler for a continuation of research efforts towards the real-life aerial cooperative inspection of aging infrastructure, a concept that has never been presented before to the authors best knowledge, in outdoor and with a real infrastructure as a test case. In the outdoors demonstrations, a case depicted in Figure~\ref{fig:hexacopter_cov}, the aerial UAVs have been autonomously operated based on odometry information from visual and inertial sensor fusion and without any other support on localization (e.g. motion capturing system), which adds more complexity and impact on the acquired results, while the image and pose data on board the platform were post processed to build a 3D representation of the structure. An additional contribution of the proposed scheme, is the fact that the proposed C-CPP is considering the coverage problem in the case of 3D structure and it is able to feed the agents with online collision-free coverage paths. In this respect, the online optimization provides multiple waypoints for each agent in order to have maximum distance between them during the inspection task.

The rest of the article is structured as follows. The proposed C-CPP method is presented in Section~\ref{Methodology}, which follows with a brief description on the 3D reconstruction from multiple agents in Section~\ref{map_merging}. In Section~\ref{results} multiple simulation and experimental results are presented. Finally the article concludes in Section~\ref{conclusion}.
%
\section{Coverage Path Planning of 3D Maps}\label{Methodology}
\begin{figure}[htbp]
\centering
\includegraphics[width=40mm]{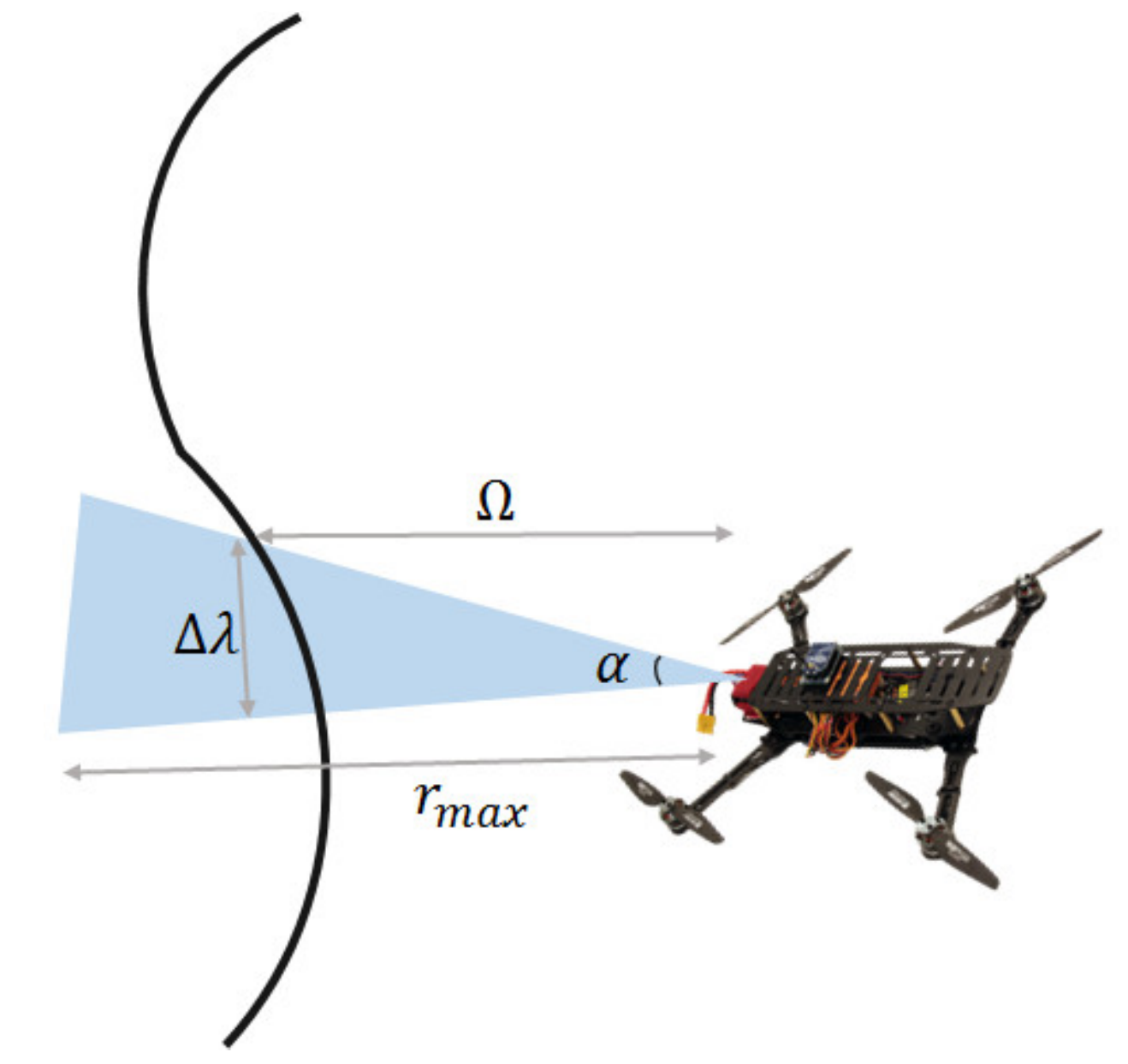}
\caption{UAV based FOV (regenerated from~\cite{galcerancoverage}).}
  \vspace{-1em}
\label{FOV}
\end{figure}

%
For the establishment of the C-CPP, initially we consider the general case of a robot equipped with a limited Field of View (FOV) sensor, determined by an aperture angle $\alpha$ and a maximum range $r_{max}$, as depicted in Figure~\ref{FOV}. Furthermore, $\Omega \in R^+$ is the user-defined offset distance ($\Omega < r_{max}$), from the infrastructure's target surface, and $\Delta \lambda$ is the distance between each slice planes and it is equal to $\frac{\Omega}{2} \tan{\alpha}$ to guarantee overlapping. 

\begin{figure}[htbp]
\centering
\includegraphics[width=85mm]{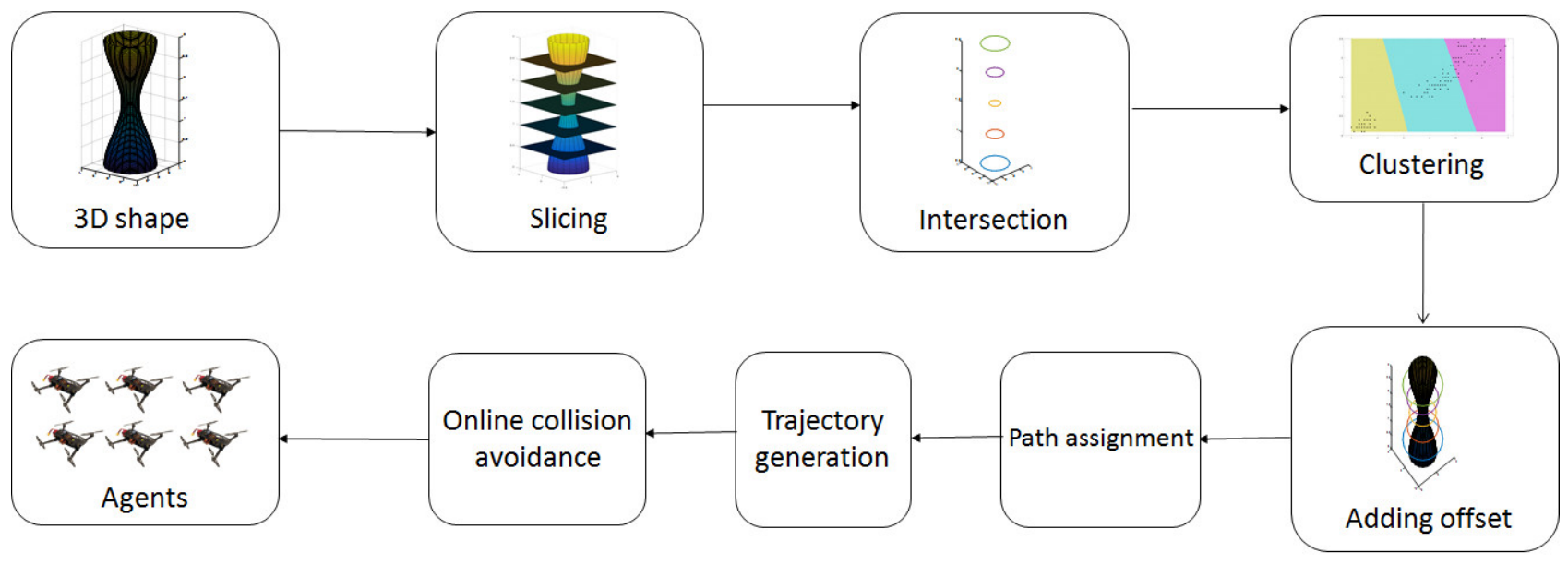}
\caption{Block Diagram of the overall proposed C-CPP scheme.}
  \vspace{-1em}
\label{MethodSchematic}
\end{figure}

The proposed C-CPP method operates offline and it assumes that a 3D shape of the infrastructure is a priori known. The execution of the proposed scheme is characterized by the following five steps: 1) slicing and intersection, 2) critical points identification, 3) adding offset from the infrastructure's surface, 4) path assignment for each UAV, 5) trajectory generation, and 6) online collision avoidance. This algorithmic approach is being depicted in Figure~\ref{MethodSchematic} and will be described in the sequel, while it should be noted that the overall scheme has been inspired from the approaches presented in~\cite{atkar2001exact} and~\cite{galceran2013planning}, implemented for only the one agent case. 
%
\subsection{Slicing and Intersection}
\begin{figure}[htbp]
\centering
\includegraphics[width=\linewidth,height=50 mm]{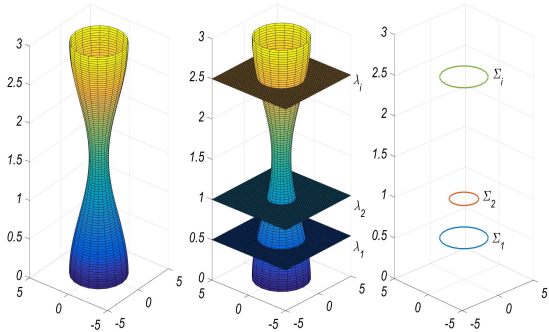}
\caption{Concept of plane slicing and intersection points.}
  \vspace{-1em}
\label{Slicing}
\end{figure}

The 3D map of the infrastructure is provided as a set $S$ of multiple points, denoted as $S(x,y,z)$, where $x,y,z \in R^3$ and in the sequel the map is sliced by multiple horizontal planes, defined as $\lambda_i$, with $i\in Z^{+}$, while the overall concept is depicted in Figure~\ref{Slicing} for the case of a wind turbine aerial inspection. The value of $\lambda_i$ initiates with a minimum of $\Delta \lambda$ and a maximum of $max_z S(x,y,z)- \Delta \lambda$, thus in this case, the horizontal plane translated vertically along the $z$-axis, while increasing the distance from the current slice, until reaching the maximum value. The intersection between the 3D shape and the slice can be calculated as it follows:
\begin{equation}
\Sigma_{i}=\{(x,y,z)\in R^3 : <n,S(x,y,z)>-\lambda_i=0\} 
\end{equation}
Where $\Sigma_{i}$ are the points of the intersection of the plane and the 3D shape, $n\in R^3$ is the vector perpendicular to the plane and $\lambda_i$ defines the location of the plane, while since in the examined case the planes are horizontal $n=[0,0,1]^T$. The slicing and intersection algorithm is presented in Algorithm \ref{Slice}, while the overall concept is depicted in Figures~\ref{Slicing} and~\ref{branch} for the case of a simple and more complex infrastructure.

\begin{algorithm}
\caption{Slicing and Intersection}
\label{Slice}
\begin{algorithmic}
\Require Points of 3D shape S(x,y,z), $\Delta \lambda$
\\
$\lambda= min_z S(x,y,z) + \Delta \lambda$ 
\While{$\lambda_i< max_z S(x,y,z)$} \\
$\Sigma_{i}=\{(x,y,z)\in R^3 :\, <n,S(x,y,z)>-\lambda_i=0\} $ \\
Clustering() ;Check Section \ref{Clustering}\\
$\lambda_{i+1}= \lambda_{i}+ \Delta \lambda$
\EndWhile
\end{algorithmic}
\end{algorithm}

%
\subsection{Critical Points}
\begin{figure}[htbp]
\centering
\includegraphics[width=50 mm,height=50 mm]{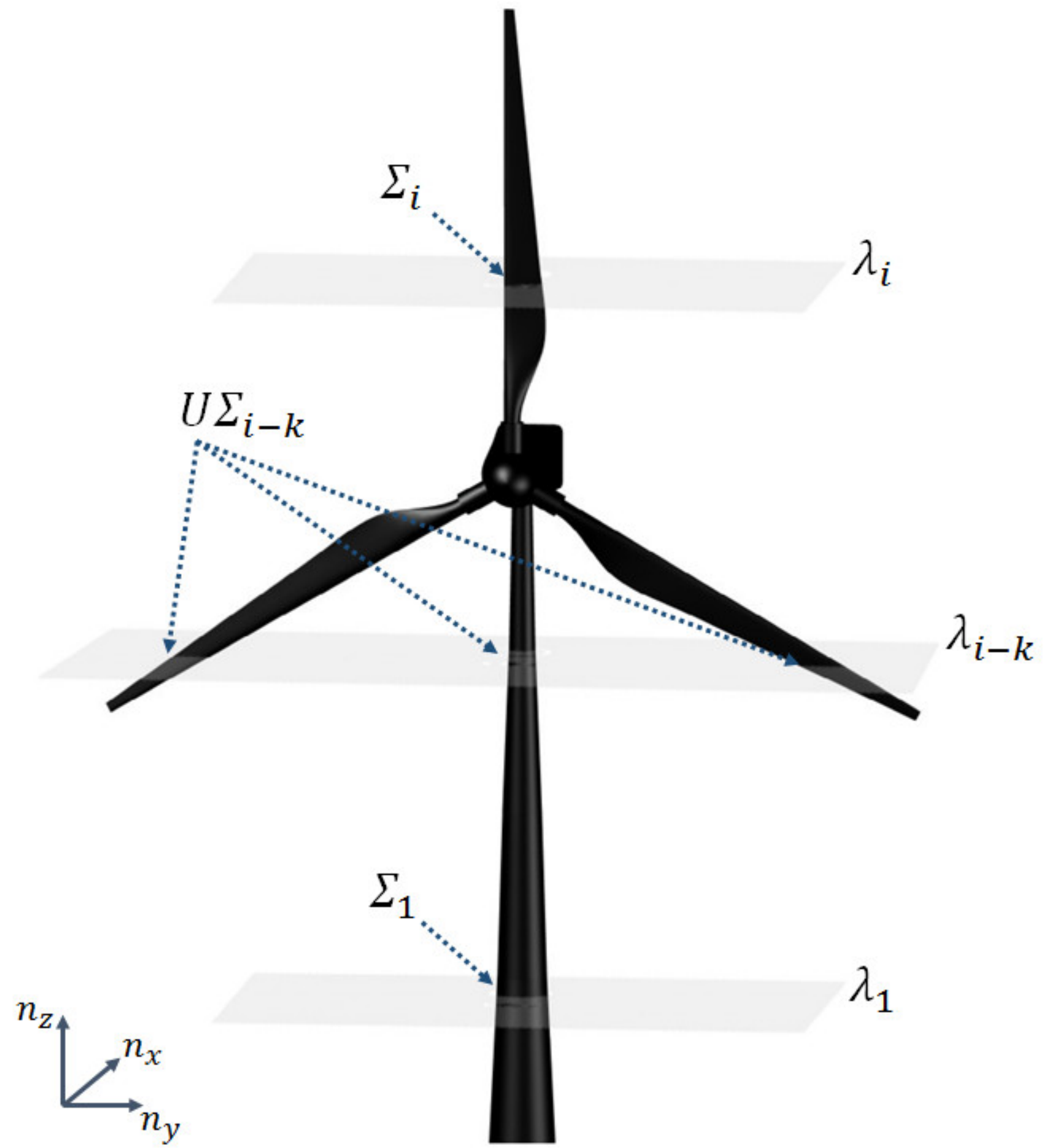}
\caption{Object with multiple branches.}
  \vspace{-1em}
\label{branch}
\end{figure}

In the case of convex infrastructures, there is at most one loop in any slice, while in the non-convex objects there might be more than one loop in each slice, as it is presented in Figure \ref{branch}. Thus, it is critical to recognize the number of the loops (number of branches in object) in order to properly complete the overall C-CPP task. 

During the execution of the slicing and intersection algorithm, the points of intersection are obtained, while the overall challenge is to determine the number of loops and separate the points that belong to the same loops. Towards the solution of this problem, two steps have been proposed for recognizing the number of loops and categorizing the points. 
\subsubsection{Number of loops} In order to obtain the number of loops graph theory will be utilized and thus initially the adjacency matrix ($A$) is generated. If the distance of the two points are less than $d_{min}$ then the points are connected, while for selecting the value of $d_{min}$, the size of the agent should be considered. Thus, if the agent cannot pass through two points, these should be assumed to be connected and merged. In the next step, the degree matrix ($D$) is generated and the Laplacian matrix ($L$) is calculated. Finally, the eigenvalues of $L$ are calculated, since the number of times that zero appears as an eigenvalue, represents the number of the connected graphs.
\subsubsection{Clustering} \label{Clustering}
The biggest challenge in using clustering algorithms is to determine the number of the clusters in a data set. While the number of connected graph is achieved by finding the number of zeros of Laplacian matrix eigenvalues, the number of connected graph has a direct relation to the number of clusters in a data set. In order to group the points the \textit{K-means} clustering algorithm is used with the a priori knowledge of the clusters's number. The \textit{K-means} clustering~\cite{lloyd1982least} is a data-partitioning algorithm that assigns $n$ sets to exactly one of $k$ ($k<n$) clusters, while minimizing the sum of the distance of each point in the cluster to the center of the cluster as described in the following C-CPP based clustering Algorithm \ref{CLUSTER}.
\begin{algorithm}
\caption{K-means proposed clustering for C-CPP}
\label{CLUSTER}
\begin{algorithmic}
\Require $\Sigma_{i}(x,y,z)$, $k$   \\
\begin{enumerate}
\item Choose $k$ initial cluster centers. 
\item calculate $\underset{\Sigma_i}{\text{min}} \,\, \sum \limits_{i=1}^k\,\sum \limits_{\Sigma_i} ||[x,y,z]^T-C_i||$
\item assign points to the closest cluster center.
\item obtain $k_{new}$ centers by computing the average of the points in each cluster 
\item repeat Steps 2 to 4 until cluster assignments do not change
\end{enumerate}
\end{algorithmic}
\end{algorithm}

The utilization of the cluster algorithm can categorize the points, independently of the order of them or the direction of slicing to the object, while providing the center of each cluster which is utilized for the calculation of the reference yaw angle for the UAVs. Lets assume that $(x_i,y_i)$ are the points in one cluster, with the center of $(x_c,y_c)$, then the reference yaw for each point is calculated by: $\psi_{r,i}=\tan^{-1}{\frac{y_i-y_c}{x_i-x_c}}$, while  the following Algorithm \ref{CLust} has been proposed for the calculation of the number of loops and the corresponding clustering. 
\begin{algorithm}
\caption{Number of loops and clustering}
\label{CLust}
\begin{algorithmic}
\Require $\Sigma_{i}$, $d_{min}$   \\
$[m,\,n,\,p]=size(\Sigma_{i})$ 
\For{$i:1:m$} 
\For{$j:1:m$}\\ 
$d=\sqrt[]{(x(i)-x(j))^2+(y(i)-y(j))^2}$
\If{$d<d_{min}$} \\
$A(i,j)=1$
\EndIf
\EndFor
\EndFor
\\
$D(i,j)=\begin{cases}
    \# ~\text{of ones in}\, i^{th} \text{row of}\, A       & \quad i==j\\
    0  & \quad  i\neq j\\
  \end{cases}$ \\
 $L=A-D$; Laplacian matrix\\ 
 $E=eig(L)$; eigen values of Laplacian matrix\\
$k$=$\# \, of \,E==0$; Number of zeros\\
 (Points, Centers)= kmeans($\Sigma_{i}$,$k$);
\end{algorithmic}
\end{algorithm}

\subsection{Adding Offset}
In the proposed C-CPP scheme, the UAV should not cover the object surface and instead it should cover the offset surface, which has a fixed distance of $\Omega$ from the target surface. Moreover, The size of the agent should also be consider in the $\Omega$. Thus, we assume that the distance between the desired points and the object is $d_i$ in the $i^{th}$ slice and the desired offset path ($OP_i$) is:
\begin{equation}
OP_{i}=\{[x,y,z]^T\in R^3: d_i-\Omega=0\}
\end{equation}
The problem of adding offset in the $R^2$ dimension is equivalent as the intersection points were on the $z$ plane. Lets assume that the $(x_i,y_i)$ should translate to the point $({x_i}^\prime,{y_i}^\prime)$ with a distance of $\Omega$. The points in Equation \ref{OffsetCirc} with the arbitrary $\theta$ have a constant distance of $\Omega$ to $(x_i,y_i)$.
\begin{equation}
\label{OffsetCirc}
\begin{aligned}
{x_i}^\prime =& x_i + \Omega cos(\theta) \\
{y_i}^\prime =& y_i + \Omega sin(\theta)
\end{aligned}
\end{equation}
The distance between two point can be calculated as follow:
\begin{equation}
\begin{aligned}
d =& \,\sqrt[]{(x_i-{x_i}^\prime)^2+(y_i-{y_i}^\prime)^2}= \\
&\, \sqrt[]{(x_i-{x_i} - \Omega cos\theta)^2+(y_i-y_i - \Omega sin\theta)^2}=\\
& |\Omega|\, \sqrt[]{sin\theta^2+cos\theta^2} =|\Omega|
\end{aligned}
\end{equation}
Moreover, in the clustering algorithm the center of each cluster is also calculated as $(x_c,y_c)$, thus the $\theta_i$ can be found for each point as it follows:
\begin{equation}
\theta_i= \tan^{-1}{\frac{y_i-y_c}{x_i-x_c}}
\end{equation}
However, as it is suggested in \cite{galcerancoverage} a more general solution for adding offset is described in \cite{liu2011fast}, where the offset surface is generated on a 3D model. In the case that \cite{liu2011fast} is used then this should be performed before the slicing and intersection step.
\subsection{Path Assignment}
For sharing the aerial infrastructure inspection to multiple UAVs two different cases have been considered, depending on if there is only one branch (loop) ($m=1$) or multiple number of branches ($m>1$). In the first case, each agent should cover part of the branch, while in the case of two agents the $\psi_i$ and $\psi_{i+1}$ has a difference of $180^\circ$. In overall if $n$ agents exist their difference between the reference yaw of $j^{th}$ and ${j+1}^{th}$ agent should be $\frac{2\pi}{n}$. In Section \ref{Clustering} the way to calculate the reference yaw is described, thus the points can be categorized to $n$ sets base on their reference yaw difference. In the case of more than one branches, the UAVs should be assigned to separate branches, where the policy of assigning each agent to each branch is shown in Algorithm \ref{policy}, while it is assumed that the distance of flight is the same for each set, as the number of points are same in each set. Finally, it should be noted that although the way points are categorized for each agent, an online collision avoidance scheme is still needed to guarantee the overall flight safety in unpredicted or faulty situations. 
\begin{algorithm}
\caption{C-CPP based assignment of UAVs to branches policy}
\label{policy}
\begin{algorithmic}
\\
Assume to have $n$ agents and $m$ branches (loops) ($n\ge m$).   
\If{$m==1$} \\
All $n$ agents go in the same slice. \\
$|\psi_i-\psi_{i+1}|=\frac{2\pi}{n}$
\EndIf
\If{$m>1$ AND $m<n$} \\
$n-m$ agents go in first branch \\
$|\psi_i-\psi_{i+1}|=\frac{2\pi}{n-m}$ \\
$m$ agents in remaining branches.
\EndIf
\If{$m==n$} \\
$n$ agents assign to $m$ branches.
\EndIf
\end{algorithmic}
\end{algorithm}
\newline 

\subsection{Trajectory Generation}
\begin{figure}[htbp]
\centering
\includegraphics[width = 60 mm, height=50 mm]{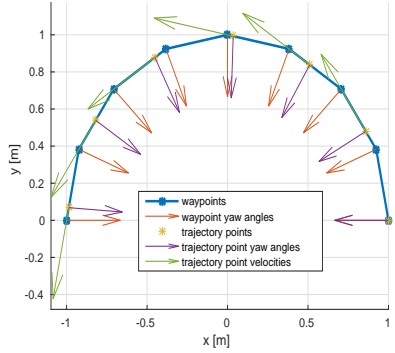}
\caption{Example application of the proposed waypoint-to-trajectory conversion algorithm.}
\label{trajectory_generation}
\end{figure}

The resulting waypoints are then converted into position-velocity-yaw trajectories, which can be directly provided to the utilized linear model predictive controller cascaded \cite{alexis2012model} over an attitude-thrust controller. This is done by taking into account the position controller's sampling time $T_s$ and the desired velocity along the path $V_d$. These trajectory points are obtained by linear interpolation between waypoints, in such a way that the distance between two consecutive trajectory points equals the step size $h = T_s \cdot V_d$. The velocities are then set parallel to each waypoint segment, with norm $V_d$, and the yaw angles are also linearly interpolated with regard to the position withing the segment. The adopted trajectory generation that was used in the experimental realization of the proposed C-CPP is depicted in Figure \ref{trajectory_generation} with $V_d = 0.5\unit{m/s}$ and $T_s = 1\,\unit{s}$.

\subsection{Collision Avoidance}
In order to avoid collisions, the following method is used to guarantee the maximum distance between the UAVs. Lets assume to have $n$ agents and that $n$ set of points are obtained for each slice (constant $z$). The first set of points ($\{(x_{1},y_{1}),(x_{2},y_{2}),\ldots,(x_{j},y_{j})\}$) will be sent to the first agent one after another. For the second agent, the following optimization is solved to find the points with the maximum distance with respect to the point allocated to the first agent. In general, the way points for the first agent are assigned without any limitation, while for the others should pass through the optimization, which means that the point of ${j+1}^{th}$ ($j\ge 1$) agent is calculated with respect to the ${j}^{th}$ agent. After the points are sent to UAVs, there are removed from the set.

The collision avoidance scheme that will be described in the sequel has been performed under the following assumptions: 1) the $\Omega$ is large enough to avoid the collision of agents to the object, 2) the distance between the branches of the object is more than the safety distance ($d_s$) thus the agents in separate branch cannot collide, and 3) the entering and leaving of the agents for each loop is a collision free path. 
\begin{equation}
\begin{aligned}
\underset{x_j,y_j}{\text{max}}&\,\, D\\
&D>d_s\\
&[x_j,y_j,z_j]^T \in \Sigma_{i} \\
&j \ge 2
\end{aligned}
\end{equation}
where $D$ is the the distance of the $j^{th}$ agent to the ${j+1}^{th}$ agent and is calculated as: 
\begin{equation}
D=\sqrt[]{(x_j-x_{j+1})^2+(y_j-y_{j+1})^2}
\end{equation}
The above optimization is \textit{Integer Linear Programming} as the optimization should find the number of sets ($\{(x_j,z_j)\}$) and the solution is the set that maximizes the distance between agents. Moreover, when the agents have to change a branch should not collide to the object as the C-CPP only provides the initial point $\{(x_s, y_s, z_s)\}$ and destination point $\{(x_d, y_d, z_d)\}$ for moving the agent from one branch to another branch. Thus, in general it is possible that the intermediate points in this path are closer than the safety distance or inside the object. For avoiding these cases, the line is produced by connecting the initial point and destination. If the points of the line are closer than the safety distance to the object, an offset value is added until the distance is larger than the safety distance. The direction of adding distance corresponds to the center of the loop as calculated before.
\section{Multiple agent Visual Inspection}\label{map_merging}

\begin{figure*}[htbp]
\centering
        \includegraphics[width = 100 mm]{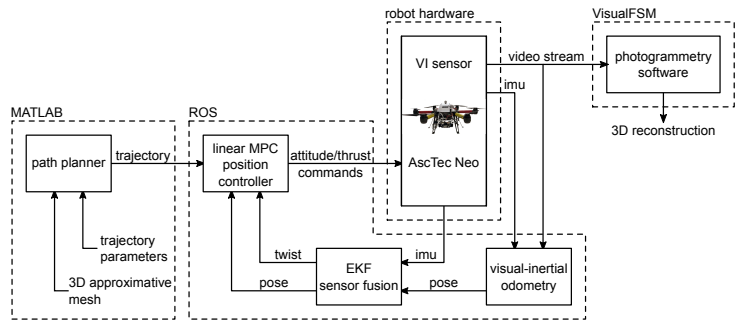}
        \caption{Software and hardware components used for conducting inspections.}
        \label{fig:experimental_setup}
    \end{figure*}

As stated throughout this article the C-CPP method is targeting the case of autonomous cooperative inspection by multiple aerial UAVs, where each of them will be equipped with a vision based system to provide a model of the 3D infrastructure, after an off line processing stage. In order to obtain the 3D model of the infrastructure, two main approaches have been utilized, which are the stereo and monocular mapping. In the stereo mapping the registration of 3D maps is achieved by using the Iterative Closest Point (ICP) \cite{BMN92} algorithm, while in the case of $n$ agents, each agent provides a unique map and in the sequel the $n$ maps are merged to form a whole structure. In the case of monocular mapping, the Structure from Motion (SfM) \cite{SF16} technique is used to merge image streams. The selection between these two approaches are mainly application dependent, however in this article both methods were used as it will be presented in Section \ref{results}.

The stereo 3D map generation is performed using the RTABMap~\cite{LM11} SLAM algorithm that is suitable for large scale operations and fits well for complex structure mapping. RTABMap is an appearance based Localization and Mapping algorithm that consists of three parts, the visual odometry and the loop closure detection, as well as the graph optimization part. In this work loop closure and graph optimization parts are needed for 3D reconstruction and Visual odometry is provided by external sources.
 
In the SfM process, different camera viewpoints are used offline to reconstruct 3D structure, a process that starts with feature extraction and matching between image frames. Afterwards, point triangulation and camera pose estimation are performed as an initial guess for the optimization step (Bundle Adjustment) to refine the reconstructed map. 



\section{Experimental Results}\label{results}
%

\begin{figure}[htbp]
    \centering
        \includegraphics[height=25mm]{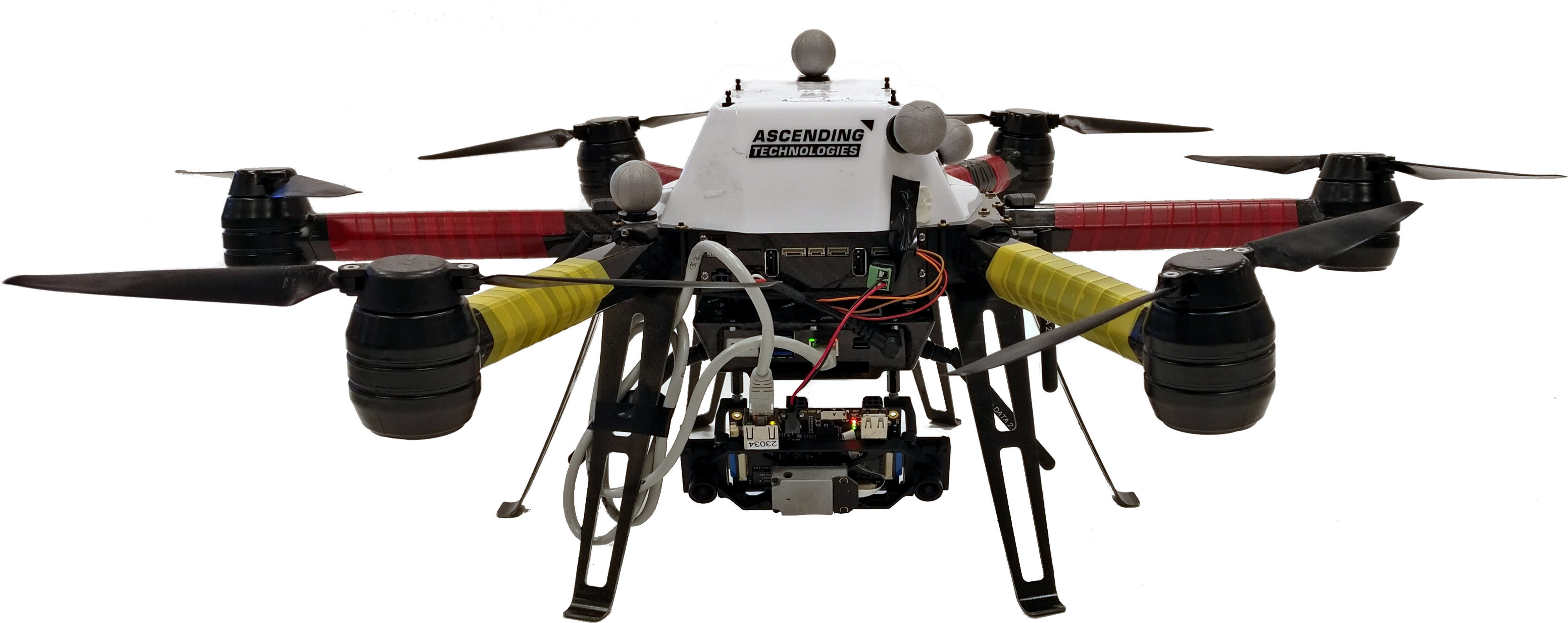}
        \caption{AscTec NEO platform with the VI sensor attached.}
        \label{fig:NEOrefVI}
    \end{figure}

The proposed method has been evaluated with the utilization of the Ascending Technologies NEO hexacopter, depicted in Figure \ref{fig:NEOrefVI}. This platform has an onboard Intel~NUC computer utilizing a Core i7-5557U combined with 8 GB of RAM and is capable of providing a flight time of 26\,$\unit{min}$. For inspection purposes, the Visual-Inertial (VI) sensor (Figure \ref{fig:NEOrefVI}) is attached below the hexacopter with a 45\,$\unit{^\circ}$ tilt from the horizontal plane. The proposed C-CPP method, established in Section \ref{Methodology}, has been entirely implemented in MATLAB, while the generated paths are sent to the NEO platforms through the utilization of the Robotic Operatic System (ROS) framework. The platform contains three main components to provide autonomous flight, which are a linear Model Predictive Control (MPC) position controller, the visual-inertial odometry and an Extended Kalman Filter (EKF) for sensor fusion and sensor bias estimation. The linear MPC position controller~\cite{mav_linear_mpc, ANT11} generates attitude and thrust references for the NEO predefined low level controller, while the visual-inertial odometry is based on the Rovio \cite{rovio} algorithm that uses data from the VI sensor and the Inertial Measurement Unit (IMU) for pose estimation. Afterwards, the EKF \cite{msf} component fuses the obtained pose information and the NEO IMU data. The image stream from the overall experiment is processed using the discussed method in Section~\ref{map_merging}, while the overall schematic of the experimental setup is presented in Figure \ref{fig:experimental_setup}.
    
\begin{figure*}[ht]
    \centering
        \includegraphics[height=60mm]{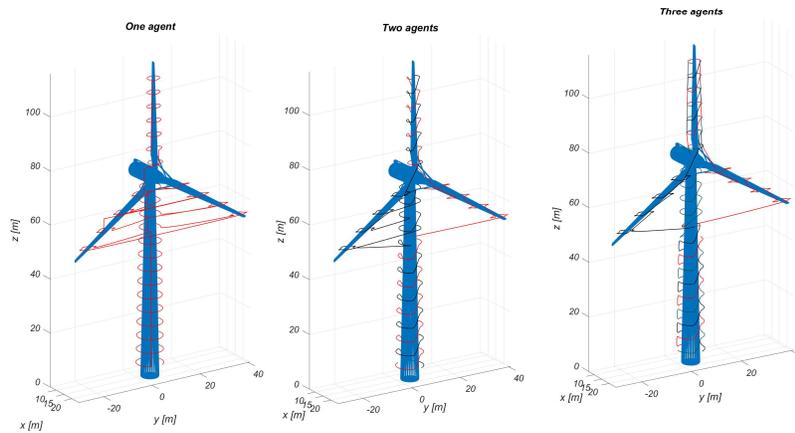}
    \caption{Generated path for different scenarios.}
    \label{fig:ALLagent}
\end{figure*}

In this section, simulation and experimental results are presented to prove the concept of the proposed method. In both cases the UAVs are equipped with visual sensors with field of view of $60^\circ$, while an offset distance $\Omega$ from the inspection objects have been adopted that varied in each case. Initially, in order to evaluate the performance of the method, a wind turbine, as a 3D complex structure, with multiple branches is selected. In Figure \ref{fig:ALLagent} the paths which are generated for one, two and three UAVs for the aerial inspection of the overall structure are depicted. In the first case (left) the structure is covered by only one agent, in the second (middle) and third (right) cases each agent is assigned to cover a specific part of the structure (branch). As a result, the inspection time is significantly reduced as presented in Table \ref{InspectionTime}.

\begin{figure}[H]
    \centering
        \includegraphics[width=\linewidth, height=60 mm]{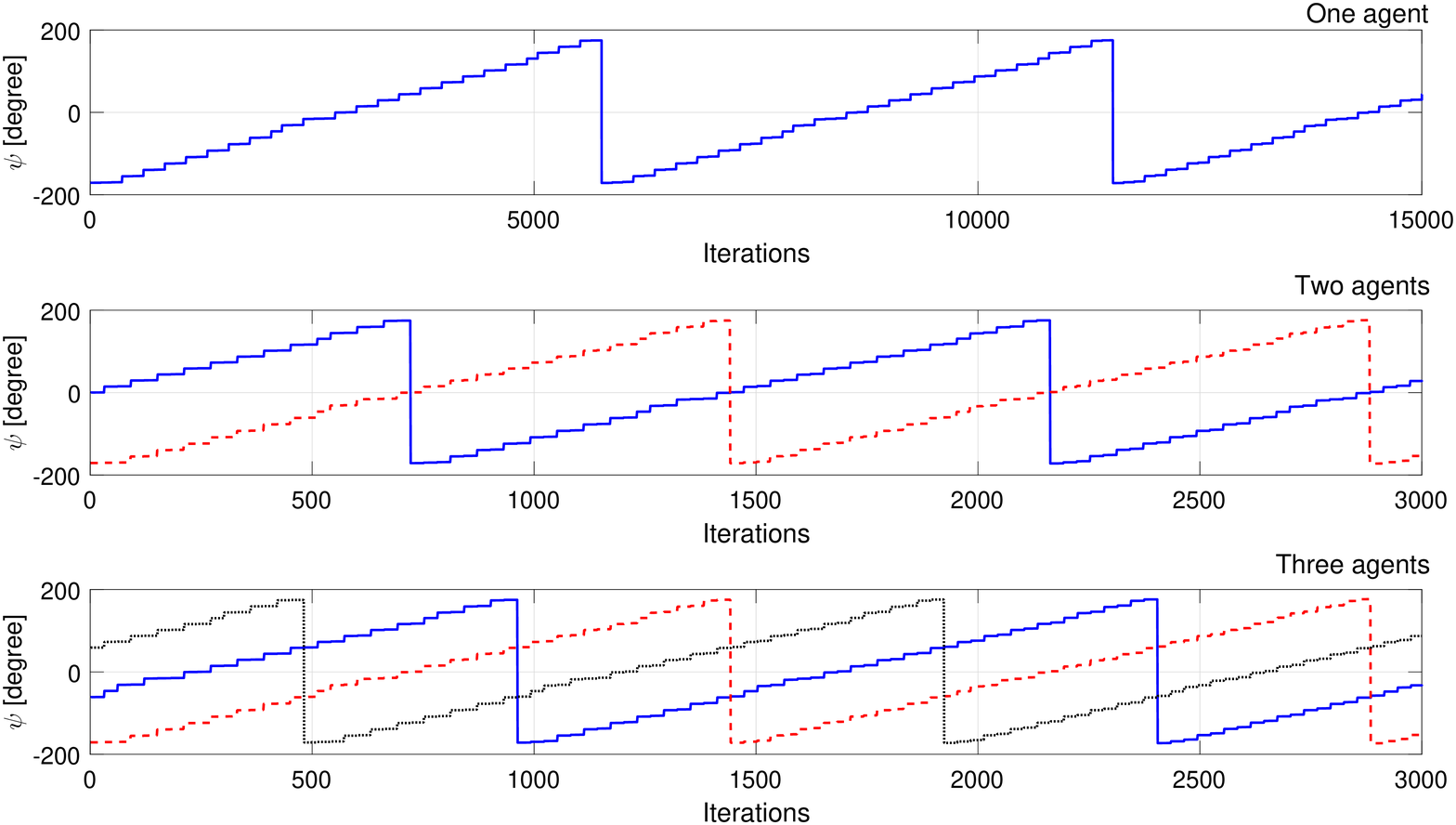}
        \caption{Yaw references for each scenario.}
        \label{fig:Yawref}
    \end{figure}

{\renewcommand{\arraystretch}{1.5}
\begin{table}[htbp]
\centering
\caption{Inspection Time}
\label{InspectionTime}
\begin{tabular}{|c|c|c|c|}
\hline
Number of agents                   & 1     & 2     & 3     \\ \hline
Inspection time {[}$\unit{min}${]} & 24.86 & 17.63 & 11.36 \\ \hline
\end{tabular}
\end{table}
}
Figure \ref{fig:Yawref} presents the yaw references, which have been provided for each agent in different scenarios based on the proposed C-CPP. As the number of agents is increasing the coverage task is completed faster thus the yaw changes more frequently as it has been indicated from the provided simulation results.

\begin{figure}[htbp]
    \centering
        \includegraphics[width=\linewidth,height=50 mm]{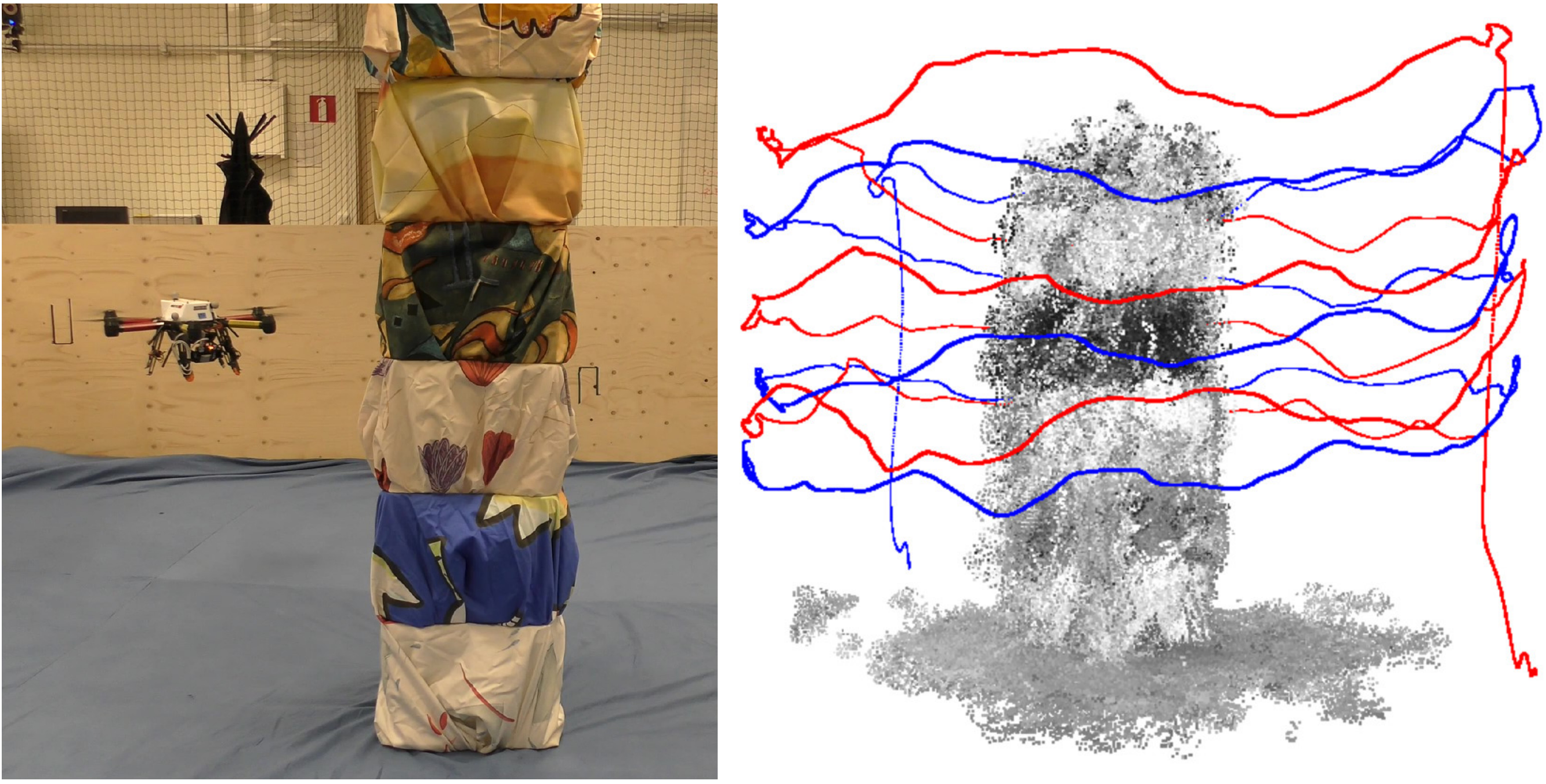}
        \caption{On the left is the simple indoor structure to be reconstructed and, on the right, the cooperative pointcloud of the structure.}
        \label{indoorStructure}
    \end{figure}

For demonstrating the applicability of the method, two inspection scenarios have been performed. For the first scenario an indoor artificial substructure was made in Lule{\aa} FROST Lab (Figure \ref{indoorStructure}). The structure consisted of $6$ boxes with dimensions of $57\times40\times30$ cm with unique patterns and without any branches. In this case, two aerial agents were assigned to cover the structure. The FROST Lab is equipped with a Vicon Motion-capture (Mo-cap) system that have been utilized for the precise object localization. In the sequel, this information has been utilized from the NEO for the autonomous flight. After the end of the experiment, the pose data from the Mo-cap system and the stereo stream have been utilized in the mapping algorithm. The actual and the reconstructed structure are showed
in Figure \ref{indoorStructure}. 
   
\begin{figure}[htbp]
    \centering
        \includegraphics[height=40 mm]{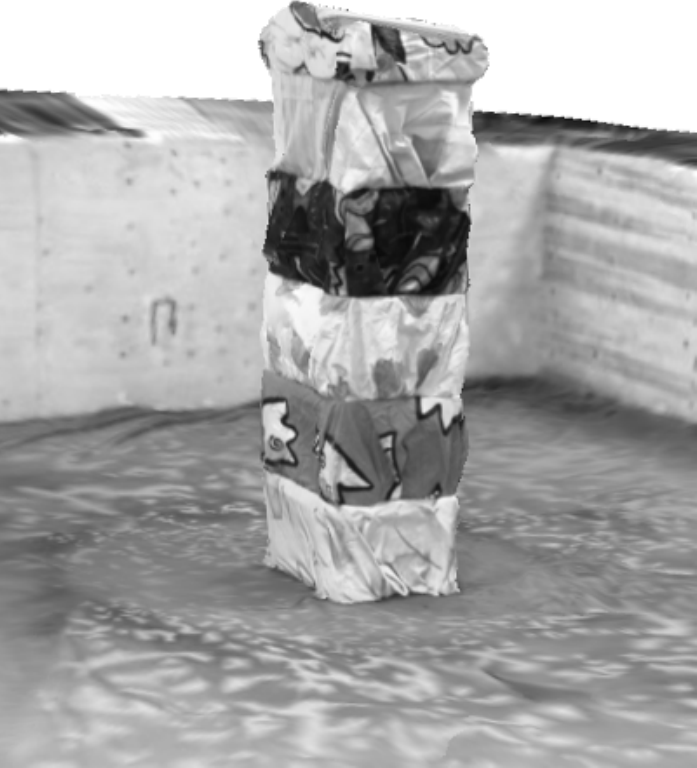}
        \caption{Cooperative 3D mesh of the indoor structure.}
        \label{indoorReCap}
    \end{figure}

\begin{figure*}[htbp]
    \centering
        \includegraphics[height=50 mm]{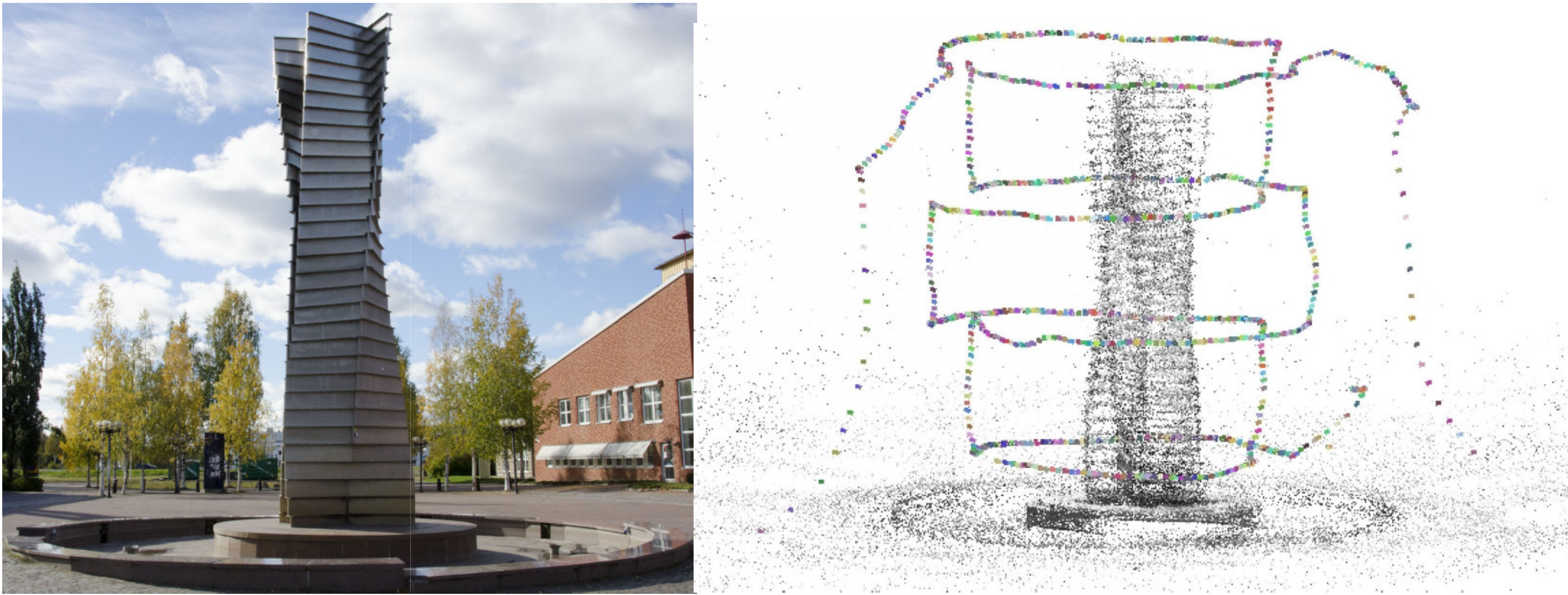}
        \caption{On the left is the Lule{\aa} University outdoor fountain, and, on the right, the cooperative pointcloud of the structure with estimated flight trajectories.}
        \label{FountainCombined}
    \end{figure*}

From the obtained results it is shown that the two agents follow the blue and red trajectories. Additionally, the starting point has 180\,$\unit{^\circ}$ difference and the agents completed the mission in 166\,$\unit{s}$ instead of 327\,$\unit{s}$. The average velocity along the path was 0.2\,$\unit{m/s}$ and the points fed to the agents in a way to avoid collision.
To retrieve the 3D mesh of the structure Autodesk ReCap 360 is used.  ReCap 360 is an online  photogrammetry software suited for accurate 3D modeling. The reconstructed surface obtained from image data, is shown in Figure \ref{indoorReCap}. 
To evaluate the performance of the method in the real autonomous inspection task, an outdoors experiment was conducted. For this purpose the Lule{\aa} University's campus fountain has been selected to represent the actual infrastructure for the cooperative aerial inspection. The fountain has a radius of 2.8\,$\unit{m}$ and height 10.1\,$\unit{m}$ without branches. Since in the outdoor experiments, motion capturing systems are rarely available, in order to achieve a full autonomous flight, the localization of the UAV relied only on the onboard sensory system. Thus, the UAVs followed the assigned paths with a complete onboard computation (fully realistic autonomous flight). For the reconstruction the image stream from both agents are combined and fed through the SfM algorithm. The fountain and the result are shown in Figure \ref{FountainCombined} and the collision free paths of both agents are shown.

\begin{figure}[htbp]
    \centering
        \includegraphics[height=50 mm]{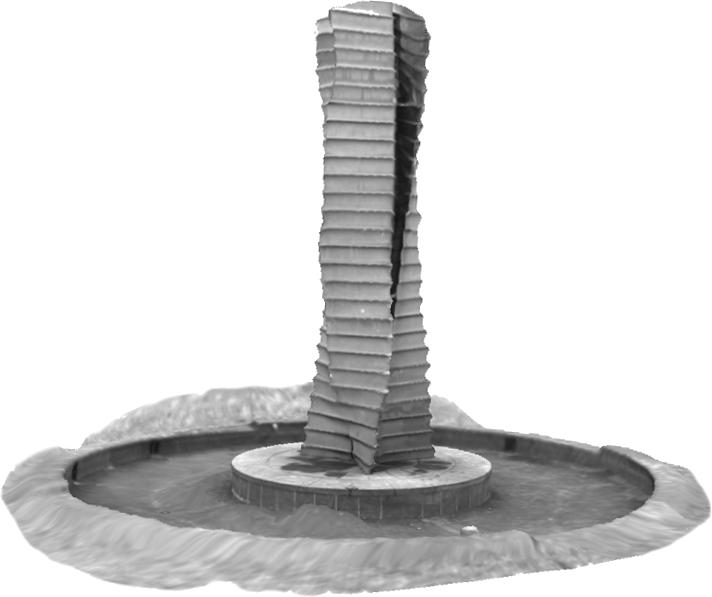}
        \caption{Cooperative 3D mesh of the outdoor structure.}
        \label{OutdoorReCap}
    \end{figure}

In the proposed experiment the same strategy as indoor experiment is followed for two agents. The starting position of each of them has the maximum of distance with 180\,$\unit{^\circ}$ difference. The overall flight time is reduced from 370\,$\unit{s}$ to 189\,$\unit{s}$ and the average velocity along the path was 0.5\,$\unit{m/s}$. 
The sparse reconstruction provided in Figure \ref{FountainCombined} cannot be used for inspection tasks, since it lacks texture information and contains noise. 
Similarly to indoor experiment the reconstructed surface obtained from image data, is shown in Figure \ref{OutdoorReCap}. The results show that the collaborative scheme of the path planner could be successfully integrated for automating inspection tasks.
     
\section{Conclusions} \label{Conclusion}\label{conclusion}
%
This article addresses the C-CPP for the inspection of complex infrastructures by utilizing multiple agents. In order to achieve a complete cooperative coverage in a short time this method uses an a drpriori 3D model of the infrastructure and generate multiple paths for UAVs. The algorithm recognizes the branches of the model and assigns the different part of the infrastructure to each agent.  The performance of the proposed C-CPP has been experimentally evaluated in multiple indoor and outdoor infrastructure inspection experiments. More complex structure with multiple branches (e.g. windmill) will be studied for cooperative inspection as a future work. 
\bibliography{mybib}
\end{document}